\documentclass[11pt,a4paper]{article}
\usepackage[hyperref]{acl2019}
\usepackage{times}
\usepackage{latexsym}

\usepackage{url}

\usepackage{multirow}
\usepackage{booktabs}
\usepackage[export]{adjustbox}
\usepackage{arydshln}
\usepackage{amsmath}
\usepackage{amssymb}
\usepackage{blindtext}
\usepackage{graphicx}
\usepackage{multirow}
\usepackage{multicol}
\usepackage[super]{nth}

\newcommand{\parheader}[1]{{\bf \smallskip \noindent #1.}}

\aclfinalcopy 

\title{What Would Elsa Do? Freezing Layers During Transformer Fine-Tuning}

\author{Jaejun Lee, Raphael Tang, \and Jimmy Lin\vspace{0.1cm}\\
David R. Cheriton School of Computer Science\\
University of Waterloo}

\date{}

\begin{document}
\maketitle
\begin{abstract}
Pretrained transformer-based language models have achieved state of the art across countless tasks in natural language processing.
These models are highly expressive, comprising at least a hundred million parameters and a dozen layers.
Recent evidence suggests that only a few of the final layers need to be fine-tuned for high quality on downstream tasks.
Naturally, a subsequent research question is, ``how many of the last layers do we need to fine-tune?''
In this paper, we precisely answer this question.
We examine two recent pretrained language models, BERT and RoBERTa, across standard tasks in textual entailment, semantic similarity, sentiment analysis, and linguistic acceptability.
We vary the number of final layers that are fine-tuned, then study the resulting change in task-specific effectiveness.
We show that only a fourth of the final layers need to be fine-tuned to achieve 90\% of the original quality.
Surprisingly, we also find that fine-tuning all layers does \textit{not} always help.
\end{abstract}

\section{Introduction}
Transformer-based pretrained language models are a battle-tested solution to a plethora of natural language processing tasks.
In this paradigm, a transformer-based language model is first trained on copious amounts of text, then fine-tuned on task-specific data.
BERT~\cite{devlin2018bert}, XLNet~\cite{Yang2019XLNetGA}, and RoBERTa~\cite{liu2019roberta} are some of the most well-known ones, representing the current state of the art in natural language inference, question answering, and sentiment classification, to list a few.
These models are extremely expressive, consisting of at least a hundred million parameters, a hundred attention heads, and a dozen layers.

An emerging line of work questions the need for such a parameter-loaded model, especially on a single downstream task.
\citet{michel2019sixteen}, for example, note that only a few attention heads need to be retained in each layer for acceptable effectiveness.
\citet{kovaleva2019revealing} find that, on many tasks, just the last few layers change the most after the fine-tuning process.
We take these observations as evidence that only the last few layers necessarily need to be fine-tuned.

The central objective of our paper is, then, to determine how many of the last layers actually need fine-tuning.
Why is this an important subject of study?
Pragmatically, a reasonable cutoff point saves computational memory across fine-tuning multiple tasks, which bolsters the effectiveness of existing parameter-saving methods~\cite{houlsby2019parameter}.
Pedagogically, understanding the relationship between the number of fine-tuned layers and the resulting model quality may guide future works in modeling.

Our research contribution is a comprehensive evaluation, across multiple pretrained transformers and datasets, of the number of final layers needed for fine-tuning.
We show that, on most tasks, we need to fine-tune only one fourth of the final layers to achieve within 10\% parity with the full model.
Surprisingly, on SST-2, a sentiment classification dataset, we find that \textit{not} fine-tuning all of the layers leads to improved quality.

\section{Background and Related Work}

\subsection{Pretrained Language Models}
In the pretrained language modeling paradigm, a language model (LM) is trained on vast amounts of text, then fine-tuned on a specific downstream task.
\citet{peters2018deep} are one of the first to successfully apply this idea, outperforming state of the art in question answering, textual entailment, and sentiment classification.
Their model, dubbed ELMo, comprises a two-layer BiLSTM pretrained on the Billion Word Corpus~\cite{chelba2014one}.

Furthering this approach with more data and improved modeling, \citet{devlin2018bert} pretrain deep 12- and 24-layer bidirectional transformers~\cite{vaswani2017attention} on the entirety of Wikipedia and BooksCorpus~\cite{zhu2015aligning}.
Their approach, called BERT, achieves state of the art across all tasks in the General Language Understanding Evaluation (GLUE) benchmark~\cite{wang2018glue}, as well as the Stanford Question Answering Dataset (\citealp{rajpurkar2016squad}).

As a result of this development, a flurry of recent papers has followed this more-data-plus-better-models principle.
Two prominent examples include XLNet~\cite{Yang2019XLNetGA} and RoBERTa~\cite{liu2019roberta}, both of which contest the present state of the art.
XLNet proposes to pretrain two-stream attention-augmented transformers on an autoregressive LM objective, instead of the original cloze and next sentence prediction (NSP) tasks from BERT.
RoBERTa primarily argues for pretraining longer, using more data, and removing the NSP task for BERT.

\subsection{Layerwise Interpretability}
The prevailing evidence in the neural network literature suggests that earlier layers extract universal features, while later ones perform task-specific modeling.
\citet{zeiler2014visualizing} visualize the per-layer activations in image classification networks, finding that the first few layers function as corner and edge detectors, and the final layers as class-specific feature extractors.
\citet{gatys2016image} demonstrate that the low- and high-level notions of content and style are separable in convolutional neural networks, with lower layers capturing content and higher layers style.

\parheader{Pretrained transformers}
In the NLP literature, similar observations have been made for pretrained language models.
\citet{clark2019does} analyze BERT's attention and observe that the bottom layers attend broadly, while the top layers capture linguistic syntax.
\citet{kovaleva2019revealing} find that the last few layers of BERT change the most after task-specific fine-tuning.
Similar to our work, \citet{houlsby2019parameter} fine-tune the top layers of BERT, as part of their baseline comparison for their model compression approach.
However, none of the studies comprehensively examine the number of necessary final layers across multiple pretrained transformers and datasets.

\section{Experimental Setup}

We conduct our experiments on NVIDIA Tesla V100 GPUs with CUDA v10.1.
We run the models from the Transformers library~(v2.1.1; \citealp{Wolf2019HuggingFacesTS}) using PyTorch v1.2.0.

\subsection{Models and Datasets}
We choose BERT~\cite{devlin2018bert} and RoBERTa~\cite{liu2019roberta} as the subjects of our study, since they represent state of the art and the same architecture.
XLNet~\cite{Yang2019XLNetGA} is another alternative; however, they use a slightly different attention structure, and our preliminary experiments encountered difficulties in reproducibility with the Transformers library.
Each model has base and large variants that contain 12 and 24 layers, respectively.
We denote them by appending the variant name as a subscript to the model name.

\begin{table}[t]
    \centering
    \small
    \setlength{\tabcolsep}{3pt}
    \scalebox{0.9}{\begin{tabular}{l l c c c}
        \toprule
        Model & Embedding & Per-Layer & Output & Total \\
        \midrule
        BERT$_\text{BASE}$ & 24M (22\%) & 7M (7\%) & 0.6M (0.5\%) & 110M \\
        RoBERTa$_\text{BASE}$ & 39M (31\%) & 7M (6\%) & 0.6M (0.5\%) & 125M \\ \midrule
        BERT$_\text{LARGE}$ & 32M (10\%) & 13M (4\%)  & 1M (0.3\%) & 335M \\
        RoBERTa$_\text{LARGE}$ & 52M (15\%) & 13M (4\%)  & 1M (0.3\%) & 355M \\
        \bottomrule
    \end{tabular}}
    \caption{Parameter statistics for the base and large variants of BERT and RoBERTa. Note that ``per-layer'' indicates the number of parameters in one intermediate layer, which is more relevant to our study.}
    \label{table:params}
\end{table}

\begin{table}[t]
    \centering
    \small
    \setlength{\tabcolsep}{2pt}
    \scalebox{0.8}{
        \begin{tabular}{lccccccccc}
            \toprule[1pt]
            \multirow{2}{*}{Model} & CoLA & SST-2 & MRPC & STS-B & QQP & MNLI & QNLI  & RTE \\
            & MCC & Acc. & $\rho$ & $\rho$ & F$_1$ & Acc. & Acc. & Acc. \\
            \midrule
            BERT$_\text{BASE}$ & 58.8 & 92.7 & 90.4 & 89.5 & 87.8 & 84.3 & 91.3 & 68.2 \\
            RoBERTa$_\text{BASE}$ & \textbf{59.9} & \textbf{94.6} & \textbf{92.8} & \textbf{90.8} & \textbf{88.8} & \textbf{87.4} & \textbf{92.7} & \textbf{78.2} \\
            \midrule
            BERT$_\text{LARGE}$ & 61.8 & 93.4 & 90.6 & 89.7 & 88.3 & 86.4 & 92.2 & 71.1 \\
            RoBERTa$_\text{LARGE}$ & \textbf{66.0} & \textbf{95.5} & \textbf{92.8} & \textbf{91.9} & \textbf{89.1} & \textbf{89.9} & \textbf{94.3} & \textbf{84.5} \\
            \bottomrule
        \end{tabular}
    }
    \caption{Reproduced results of BERT and RoBERTa on the development sets.}
    \label{table:baseline}
\end{table}

\begin{table*}[!t]
    \centering
    \setlength{\tabcolsep}{3.5pt}
    \scalebox{0.96}{
        \begin{tabular}{lcccccccccc}
            \toprule[1pt]
            \multirow{2}{*}{Model} & Frozen & CoLA & SST-2 & MRPC & STS-B & QQP & MNLI & MNLI-mm & QNLI  & RTE \\
            & up to & MCC & Acc. & F$_1$ & $\rho$ & F$_1$ & Acc. & Acc. & Acc. & Acc. \\
            \midrule
            \multirow{3}{*}{BERT$_\text{BASE}$}
            & \nth{0} & 58.3 & 92.7 & 90.3 & 88.8 & 87.9 & 84.2 & 84.8 & 91.4 & 67.6 \\
            & \nth{9} & 47.5 & 90.8 & 85.4 & 88.0 & 85.3 & 82.0 & 82.4 & 89.5 & 62.3 \\
            & \nth{12} & 29.4 & 84.9 & 81.5 & 78.1 & 72.0 & 56.4 & 57.1 & 74.5 & 57.5 \\
            \bottomrule[1pt]
        \end{tabular}
    }
    \caption{Development set results of BERT, with none, some, and all of the nonoutput layer weights fine-tuned. Results are averaged across five runs.}
    \label{table:finetune}
\end{table*}

\begin{table}[t]
    \centering
    \setlength{\tabcolsep}{2pt}
    \scalebox{0.88}{
        \begin{tabular}{lcccccccccc}
            \toprule[1pt]
            \multirow{2}{*}{Model} & Frozen & CoLA & SST-2 & MRPC & STS-B \\
            &  up to & MCC & Acc. & F$_1$ & $\rho$ \\
            \midrule
            \multirow{3}{*}{BERT$_\text{BASE}$}
            & \nth{0} & 58.3 & 92.7 & 90.3 & 88.9 \\
            & \nth{9} & 47.5 & 90.8 & 85.4 & 88.0 \\
            & \nth{12} & 29.4 & 84.9 & 81.5 & 78.1 \\
            \midrule
            \multirow{3}{*}{RoBERTa$_\text{BASE}$}
            & \nth{0} & 59.4 & 94.3 & 92.3 & 90.6 \\
            & \nth{7} & 58.6 & 93.3 & 89.5 & 87.7 \\
            & \nth{12} & 0.0 & 80.2 & 81.2 & 20.0 \\
            \bottomrule[1pt]
        \end{tabular}
    }
    \caption{Development set results of all base models, with none, some, and all of the nonoutput layer weights fine-tuned. Results are averaged across five runs.}
    \label{table:finetune-all}
\end{table}

Within each variant, the two models display slight variability in parameter count---110 and 125 million in the base variant, and 335 and 355 in the large one.
These differences are mostly attributed to RoBERTa using many more embedding parameters---exactly 63\% more for both variants.
For in-depth, layerwise statistics, see Table~\ref{table:params}.

For our datasets, we use the GLUE benchmark, which comprises the tasks in natural language inference, sentiment classification, linguistic acceptability, and semantic similarity.
Specifically, for natural language inference (NLI), it provides the Multigenre NLI~(MNLI; \citealp{williams2018broad}), Question NLI~(QNLI; \citealp{wang2018glue}), Recognizing Textual Entailment~(RTE; \citealp{bentivogli2009fifth}), and Winograd NLI~\cite{levesque2012winograd} datasets.
For semantic textual similarity and paraphrasing, it contains the Microsoft Research Paraphrase Corpus~(MRPC; \citealp{dolan2005automatically}), the Semantic Textual Similarity Benchmark~(STS-B; \citealp{cer2017semeval}), and Quora Question Pairs~(QQP; \citealp{qqp}).
Finally, its single-sentence tasks consist of the binary-polarity Stanford Sentiment Treebank~(SST-2; \citealp{socher2013recursive}) and the Corpus of Linguistic Acceptability~(CoLA; \citealp{warstadt2018neural}).

\subsection{Fine-Tuning Procedure}

Our fine-tuning procedure closely resembles those of BERT and RoBERTa.
We choose the Adam optimizer~\cite{kingma2014adam} with a batch size of 16 and fine-tune BERT for 3 epochs and RoBERTa for 10, following the original papers.
For hyperparameter tuning, the best learning rate is different for each task, and all of the original authors choose one between $1 \times 10^{-5}$ and $5 \times 10^{-5}$;~thus, we perform line search over the interval with a step size of $1 \times 10^{-5}$. We report the best results in Table~\ref{table:baseline}.

On each model, we freeze the embeddings and the weights of the first $N$ layers, then fine-tune the rest using the best hyperparameters of the full model.
Specifically, if $L$ is the number of layers, we explore $N = \frac{L}{2}, \frac{L}{2} + 1, \dots, L$.
Due to computational limitations, we set half as the cutoff point.
Additionally, we restrict our comprehensive all-datasets exploration to the base variant of BERT, since the large model variants and RoBERTa are much more computationally intensive.
On the smaller CoLA, SST-2, MRPC, and STS-B datasets, we comprehensively evaluate both models.
These choices do not substantially affect our analysis.

\begin{table}[t]
    \centering
    \setlength{\tabcolsep}{1.75pt}
    \scalebox{0.88}{
        \begin{tabular}{lcccccccccc}
            \toprule[1pt]
            \multirow{2}{*}{Model} & Frozen & CoLA & SST-2 & MRPC & STS-B \\
            & up to & MCC & Acc. & F$_1$ & $\rho$ \\
            \midrule
            \multirow{3}{*}{BERT$_\text{LARGE}$}
            & \nth{0} & 61.9 & 93.4 & 90.3 & 89.8 \\
            & \nth{18} & 51.6 & 92.7 & 85.4 & 88.0 \\
            & \nth{24} & 24.4 & 87.8 & 81.3 & 71.7 \\
            \midrule
            \multirow{3}{*}{RoBERTa$_\text{LARGE}$}
            & \nth{0} & 66.1 & 95.1 & 92.2 & 92.0 \\
            & \nth{17} & 60.5 & 95.1 & 91.3 & 89.6 \\
            & \nth{24} & 0.0 & 79.2 & 81.2 & 11.2 \\
            \bottomrule[1pt]
        \end{tabular}
    }
    \caption{Development set results of all large models, with none, some, and all of the nonoutput layer weights fine-tuned. Results are averaged across five runs.}
    \label{table:finetune-all-large}
\end{table}

\begin{figure*}[ht!]
  \centering
  \includegraphics[scale=0.23]{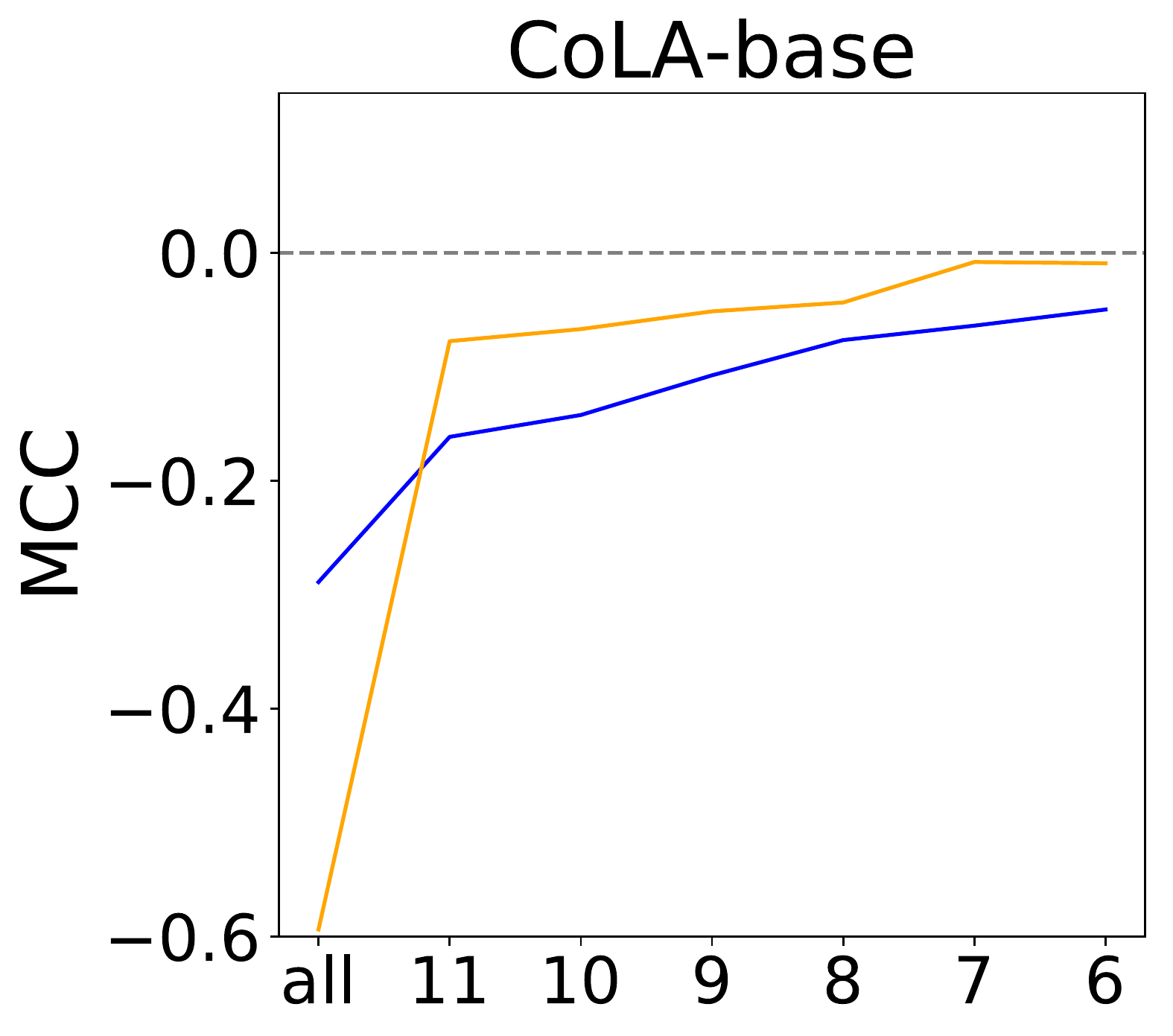}
  \includegraphics[scale=0.23]{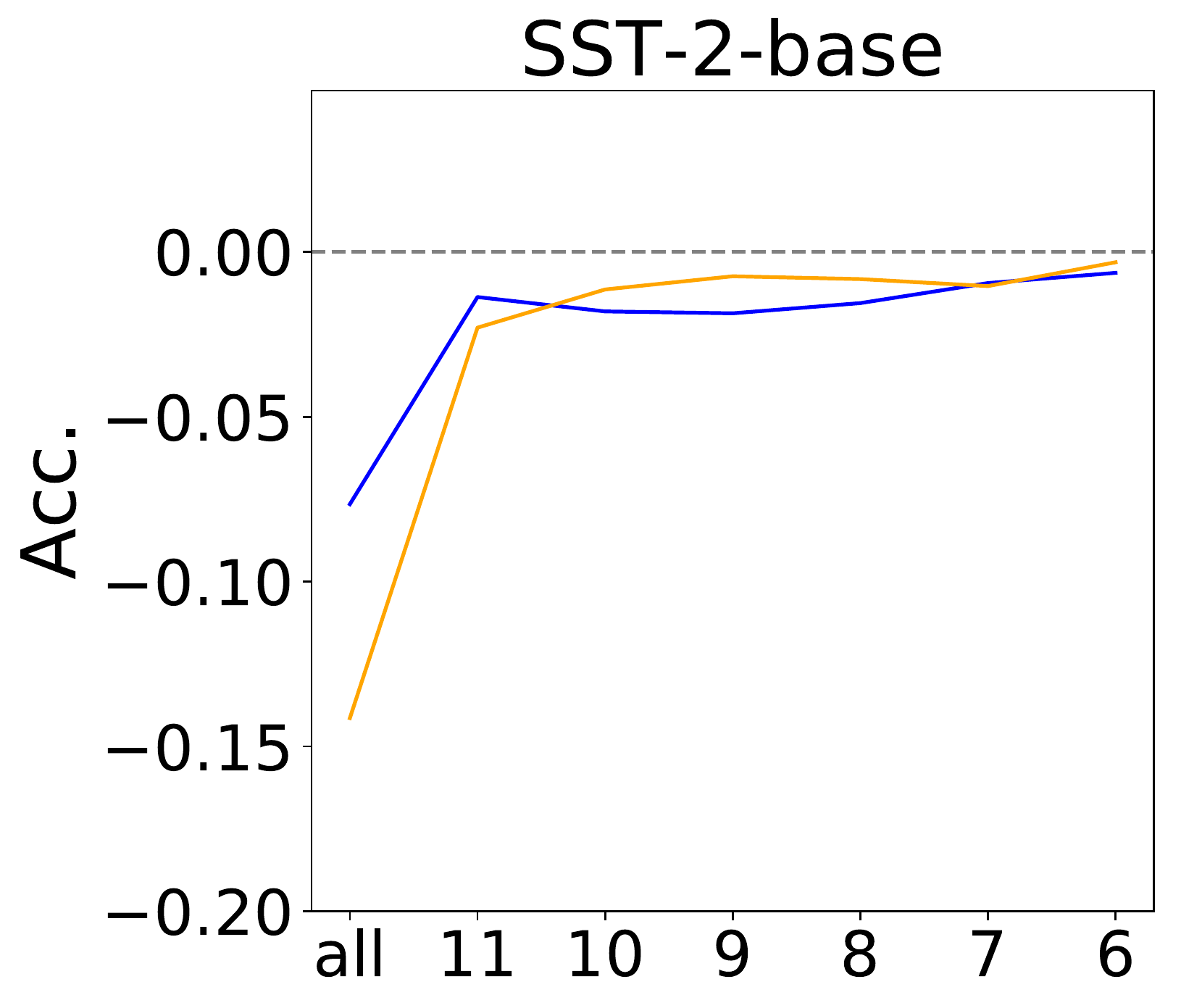}
  \includegraphics[scale=0.23]{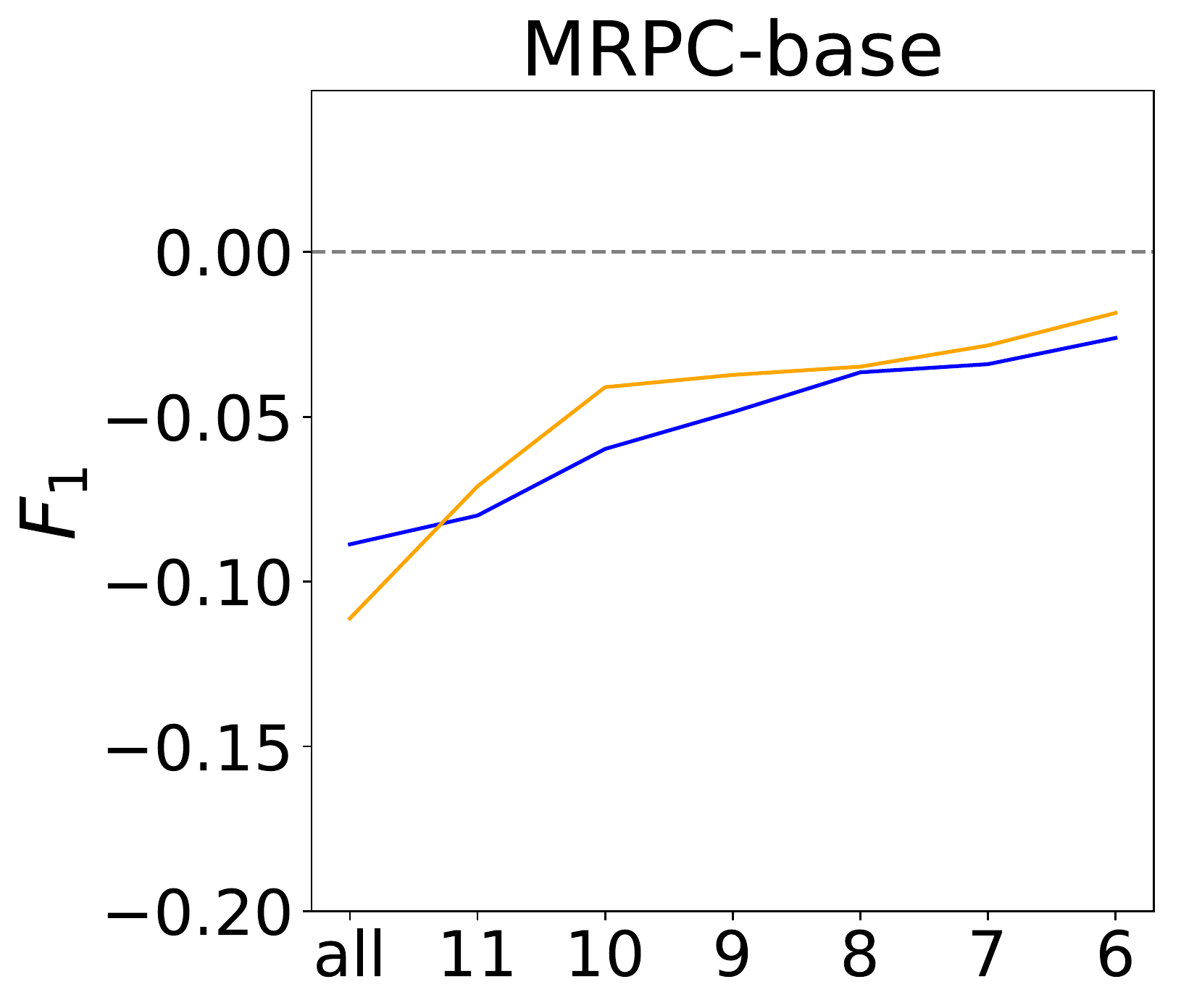}
  \includegraphics[scale=0.23]{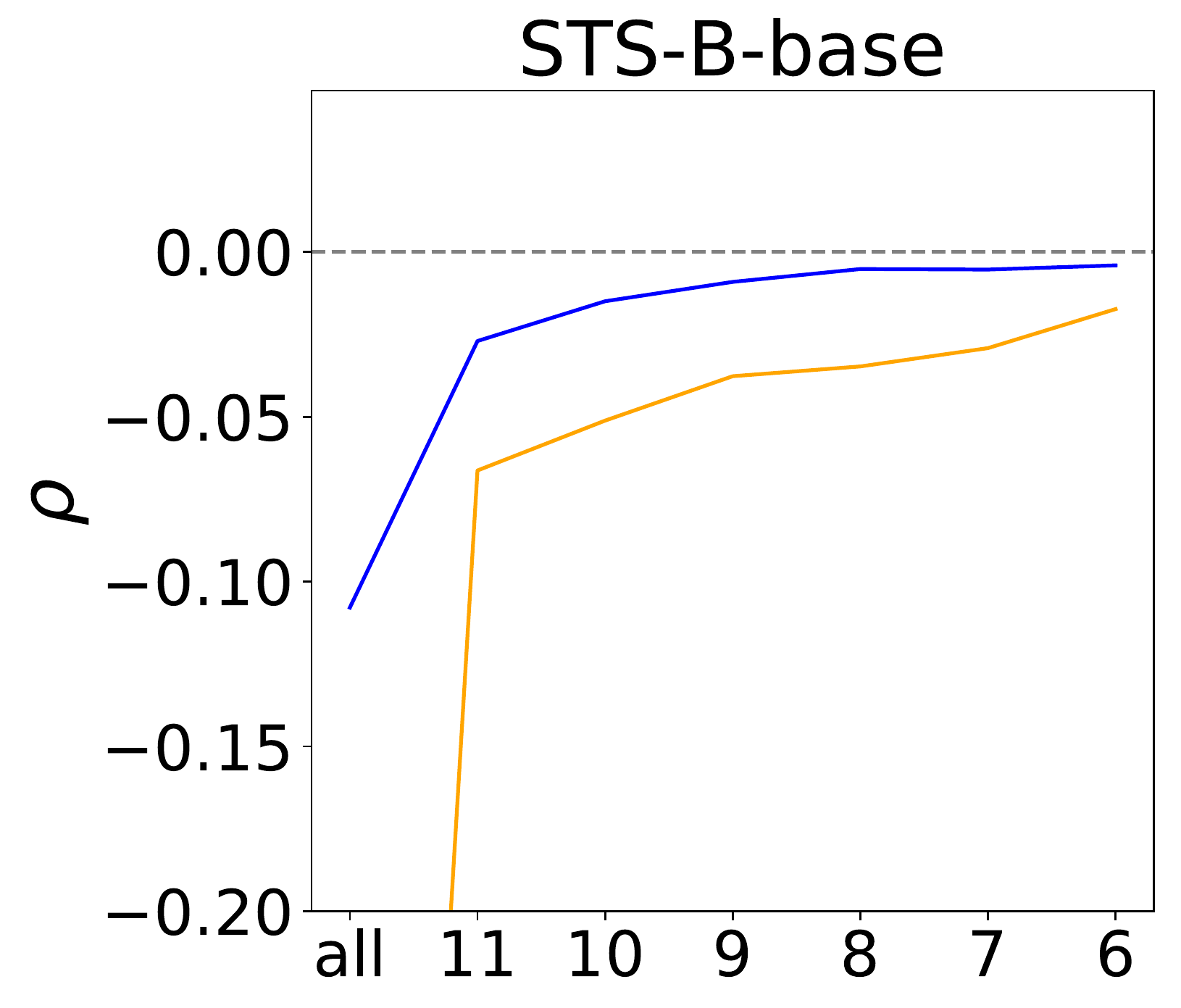}
  \\
  \includegraphics[scale=0.23]{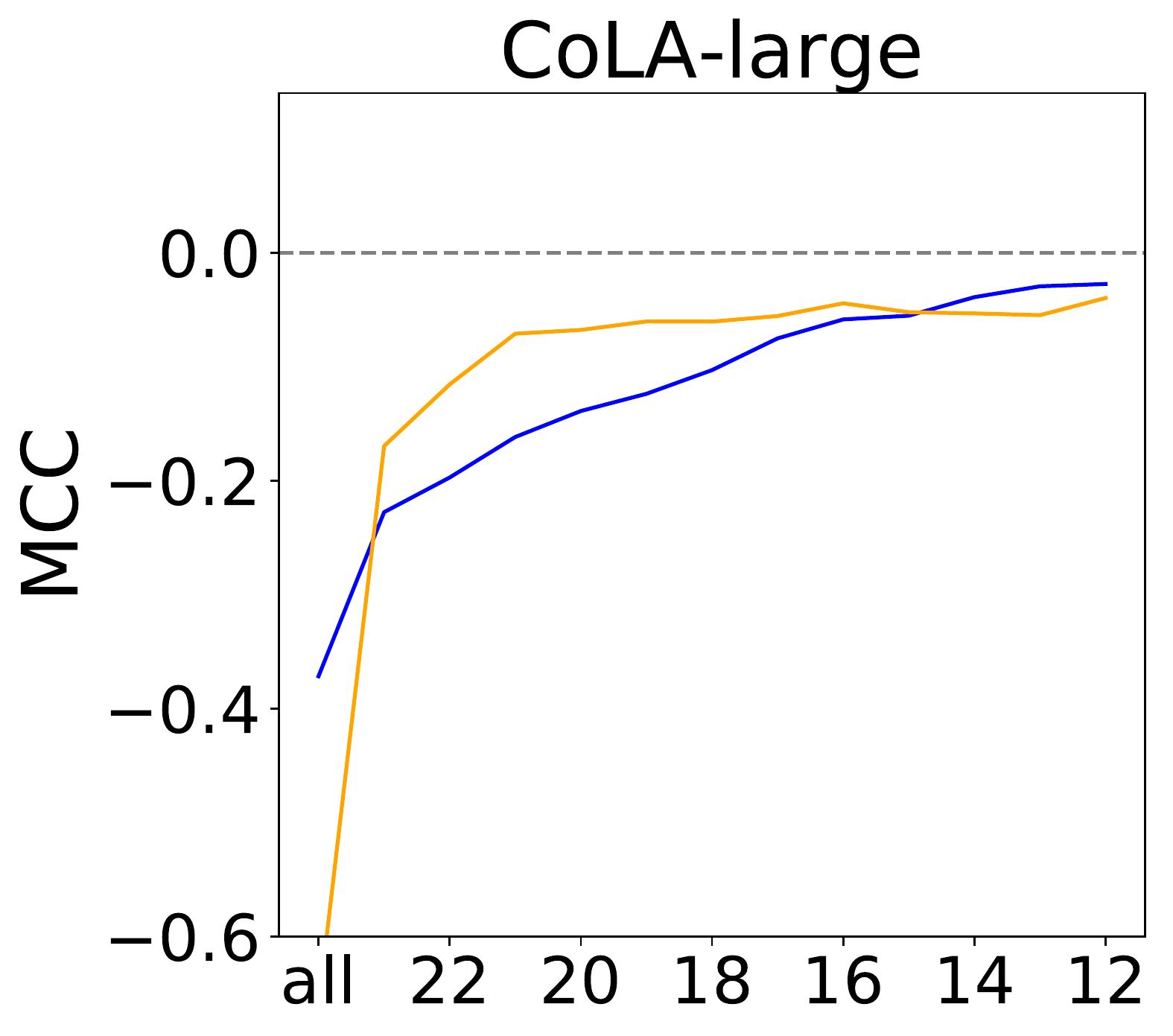}
  \includegraphics[scale=0.23]{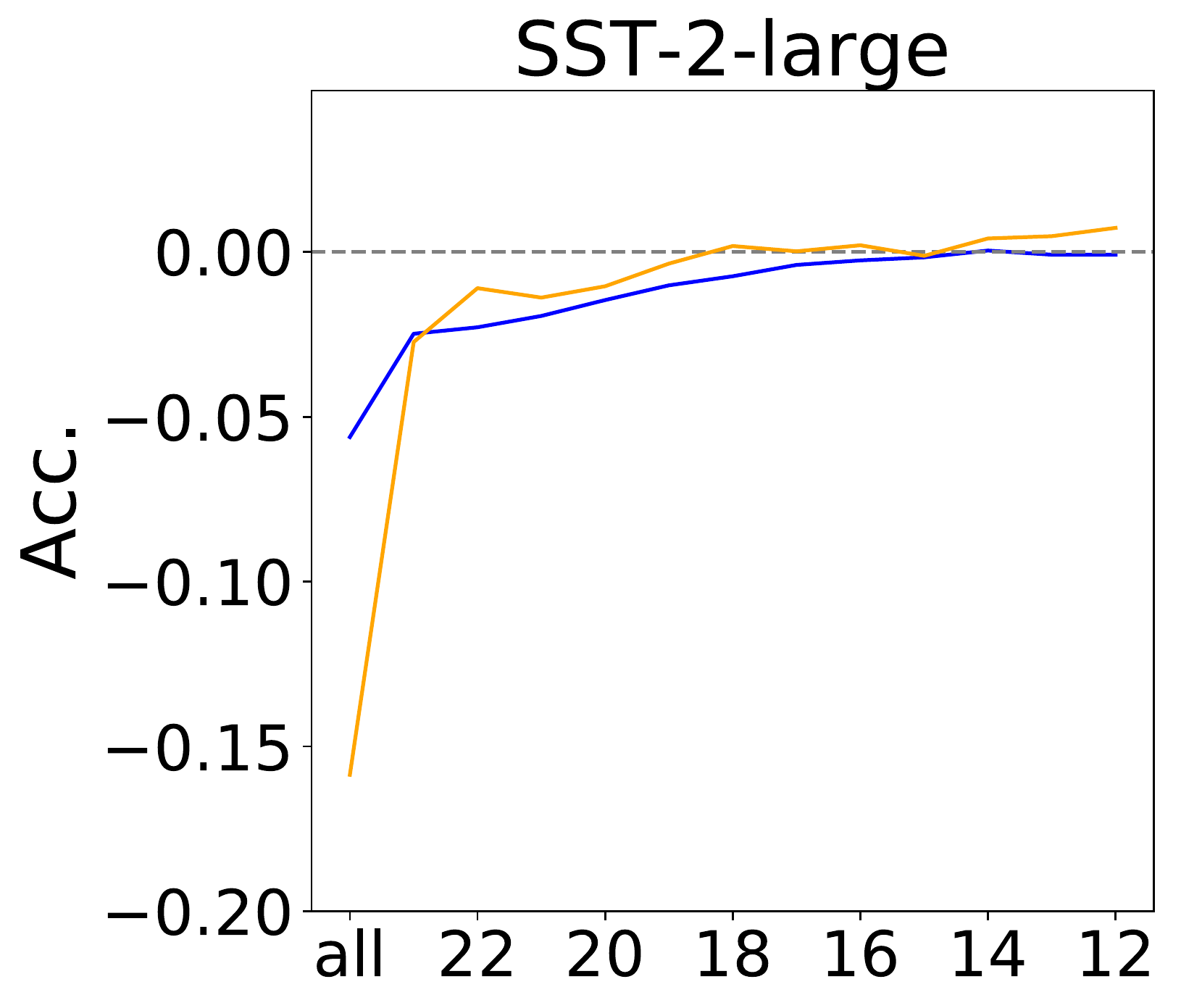}
  \includegraphics[scale=0.23]{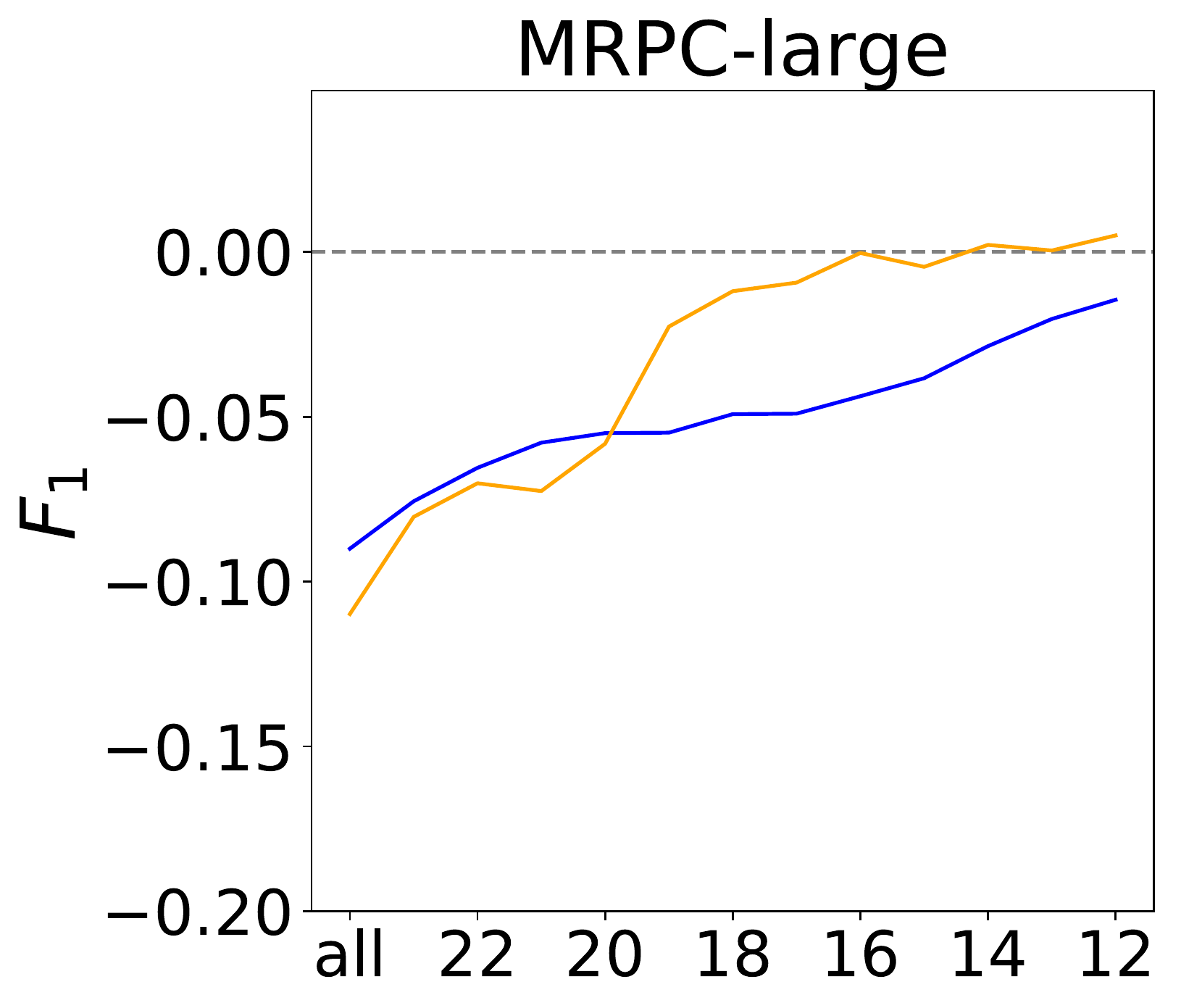}
  \includegraphics[scale=0.23]{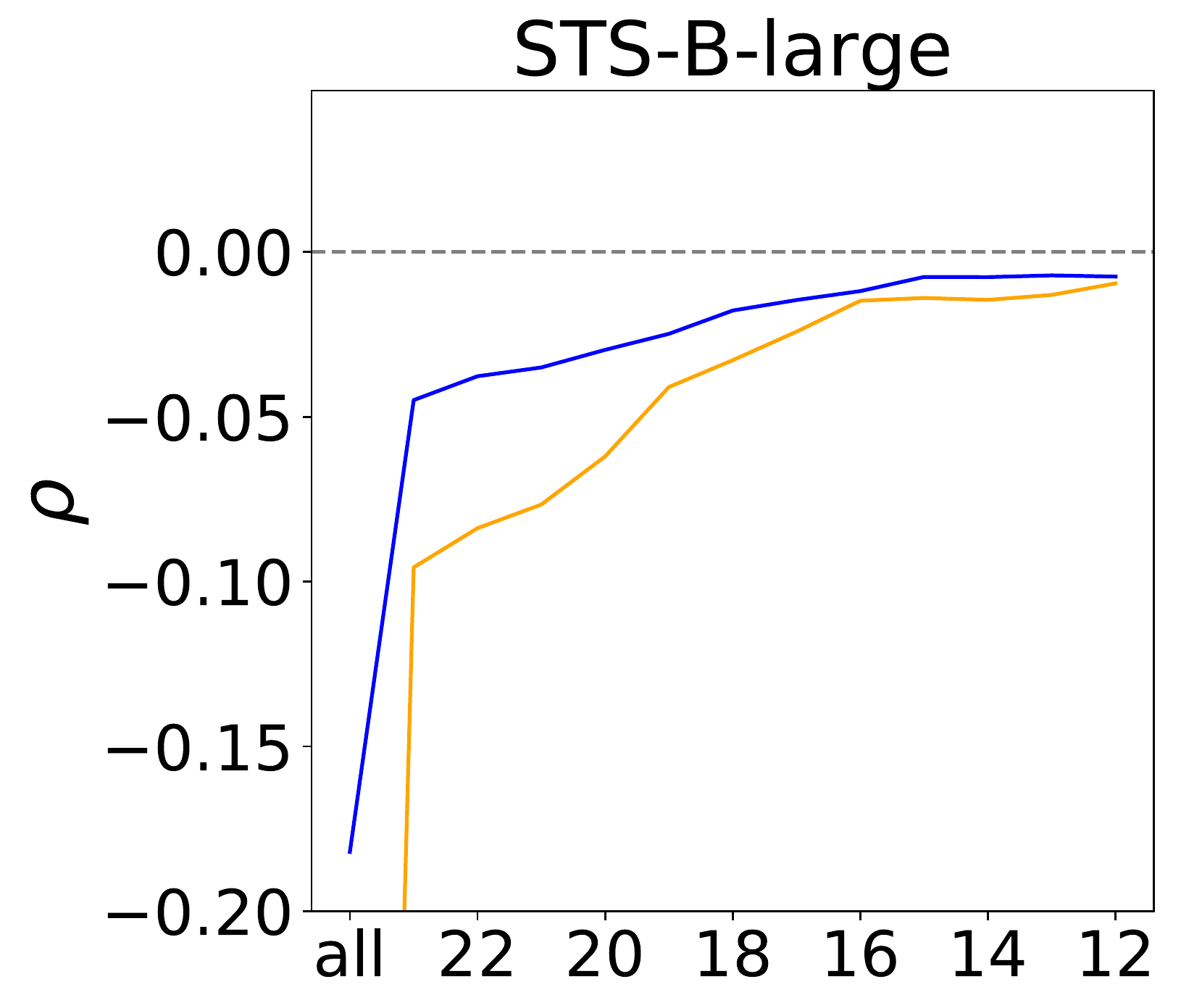}
  \\
  \includegraphics[scale=0.3]{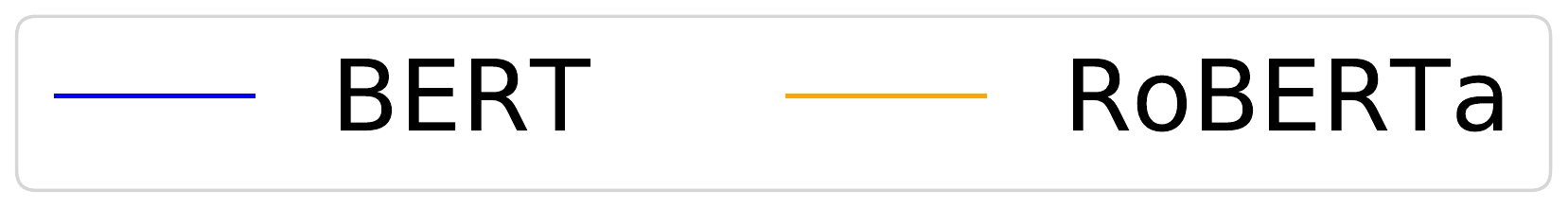}
  \caption{Relative change in quality compared to the full models, with respect to the number of frozen initial layers, represented by the $x$-axes.}
  \label{fig:diff}
\end{figure*}

\section{Analysis}
\subsection{Operating Points}
We report three relevant operating points in Tables \ref{table:finetune}--\ref{table:finetune-all-large}:~two extreme operating points and an intermediate one.
The former is self-explanatory, indicating fine-tuning all or none of the nonoutput layers.
The latter denotes the number of necessary layers for reaching at least 90\% of the full model quality, excluding CoLA, which is an outlier.

From the reported results in Tables \ref{table:finetune}--\ref{table:finetune-all-large}, fine-tuning the last output layer and task-specific layers is insufficient for all tasks---see the rows corresponding to 0, 12, and 24 frozen layers.
However, we find that the first half of the model is unnecessary;~the base models, for example, need fine-tuning of only 3--5 layers out of the 12 to reach 90\% of the original quality---see Table \ref{table:finetune-all}, middle subrow of each row group.
Similarly, fine-tuning only a fourth of the layers is sufficient for the large models (see Table \ref{table:finetune-all-large}); only 6 layers out of 24 for BERT and 7 for RoBERTa.

\subsection{Per-Layer Study}
In Figure \ref{fig:diff}, we examine how the relative quality changes with the number of frozen layers.
To compute a relative score, we subtract each frozen model's results from its corresponding full model.
The relative score aligns the two baselines at zero, allowing the fair comparison of the transformers.
The graphs report the average of five trials to reduce the effects of outliers.

When every component except the output layer and the task-specific layer is frozen, the fine-tuned model achieves only 64\% of the original quality, on average.
As more layers are fine-tuned, the model effectiveness often improves drastically---see CoLA and STS-B, the first and fourth vertical pairs of subfigures from the left.
This demonstrates that gains decompose nonadditively with respect to the number of frozen initial layers.
Fine-tuning subsequent layers shows diminishing returns, with every model rapidly approaching the baseline quality at fine-tuning half of the network; hence, we believe that half is a reasonable cutoff point for characterizing the models.

Finally, for the large variants of BERT and RoBERTa on SST-2 (second subfigure from both the top and the left), we observe a surprisingly consistent increase in quality when freezing \mbox{12--16} layers.
This finding suggests that these models may be overparameterized for SST-2.

\section{Conclusions and Future Work}
In this paper, we present a comprehensive evaluation of the number of final layers that need to be fine-tuned for pretrained transformer-based language models.
We find that only a fourth of the layers necessarily need to be fine-tuned to obtain 90\% of the original quality.
One line of future work is to conduct a similar, more fine-grained analysis on the contributions of the attention heads.

\section*{Acknowledgments}

This research was supported by the Natural Sciences and Engineering Research Council (NSERC) of Canada, and enabled by computational resources provided by Compute Ontario and Compute Canada.
\bibliographystyle{acl_natbib}

\end{document}